%% file: arxiv.tex
\documentclass[11pt,twoside]{article}

\usepackage{fullpage}
\usepackage{times}
\usepackage{bm}
\usepackage{enumitem}
\usepackage{tcolorbox}
\usepackage{amssymb}
\usepackage{subcaption,booktabs}
\usepackage[round]{natbib}

\input{arxiv_macros}

\begin{document}

\begin{center}

  {\bf{\LARGE{Partial Identification with Noisy Covariates: \\A Robust Optimization Approach}}}

\vspace*{.5in}

{\large{
\begin{tabular}{c}
Wenshuo Guo$^{\dagger}$, Mingzhang Yin$^{\diamond}$, Yixin Wang$^{\star}$, Michael I. Jordan$^{\dagger, \ddag}$
\end{tabular}
}}
\vspace*{.3in}

\begin{tabular}{c}
$^\dagger$Department of Electrical Engineering and Computer Sciences, University of California, Berkeley\\ 
$^\diamond$ Data Science Institute, Columbia University\\
$^\star$ Department of Statistics, University of Michigan\\
$^\ddag$Department of Statistics, University of California, Berkeley\\
\end{tabular}

\vspace*{.2in}

\today

\vspace*{.2in}
\end{center}

\begin{abstract}
\input{sec/sec_abstract}

\end{abstract}

\input{sec/sec_intro}
\input{sec/sec_related_work}
\input{sec/sec_prelim}

\input{sec/sec_methods-main}
\input{sec/sec_examples}
\input{sec/sec_experiment}
\input{sec/sec_conclu}

\subsubsection*{Acknowledgements}
The authors would like to thank Peng Ding for extensive discussions and helpful suggestions that significantly improved the paper. The authors also thank Peter Bickel, Avi Feller, Sam Pimentel, Vira Semenova, and Yan Shuo Tan for helpful feedback on early versions of the paper. 
This work was supported in part by the Mathematical Data Science program of the Office of Naval Research under grant number N00014-18-1-2764. WG acknowledges support from a Google PhD fellowship; MY acknowledges support from the Irving Institute for Cancer Dynamics.

\bibliography{paper-ref}
\bibliographystyle{plainnat}

\appendix

\clearpage
{\onecolumn
\input{sec/app_dro}
\input{sec/app_double_ml}
\input{sec/app_exp}
}

\end{document}

%% file: arxiv_macros.tex
\usepackage{mwe}

\usepackage{xcolor}
\definecolor{DarkGreen}{rgb}{0.1,0.5,0.1}
\definecolor{DarkRed}{rgb}{0.5,0.1,0.1}
\definecolor{DarkBlue}{rgb}{0.1,0.1,0.5}
\definecolor{Gray}{rgb}{0.2,0.2,0.2}

\definecolor{c1}{RGB}{38, 70, 83}
\definecolor{c2}{RGB}{42, 157, 143}
\definecolor{c3}{RGB}{233, 196, 106}
\definecolor{c5}{RGB}{231, 111, 81}
\definecolor{c4}{RGB}{244, 162, 97}

\definecolor{c1}{RGB}{38, 70, 83}
\definecolor{c2}{RGB}{42, 157, 143}
\definecolor{c3}{RGB}{233, 196, 106}
\definecolor{c5}{RGB}{231, 111, 81}
\definecolor{c4}{RGB}{244, 162, 97}

\usepackage{listings}
\lstdefinestyle{mystyle}{
    commentstyle=\color{DarkBlue},
    keywordstyle=\color{DarkRed},
    numberstyle=\tiny\color{Gray},
    stringstyle=\color{DarkGreen},
    basicstyle=\footnotesize,
    breakatwhitespace=false,         
    breaklines=true,                 
    captionpos=b,                    
    keepspaces=true,                 
    numbers=left,                    
    numbersep=5pt,                  
    showspaces=false,                
    showstringspaces=false,
    showtabs=false,                  
    tabsize=2
}
\usepackage{caption}
\usepackage{hyperref}
\hypersetup{
    unicode=false,          
    pdftoolbar=true,        
    pdfmenubar=true,        
    pdffitwindow=false,      
    pdfnewwindow=true,      
    colorlinks=true,       
    linkcolor=DarkBlue,          
    citecolor=DarkGreen,        
    filecolor=DarkRed,      
    urlcolor=DarkBlue,          
    pdftitle={},
    pdfauthor={},
}

\usepackage{bibunit}
\usepackage{siunitx}
\usepackage{booktabs}
\usepackage{amsfonts}
\usepackage{dsfont}
\usepackage{algorithm}
\usepackage{algorithmic}
\usepackage{mathtools}
\usepackage{natbib}
\usepackage{epstopdf}

\newcommand{\E}{\mathbb{E}}

\newcommand{\Ind}{\mathds{1}}

\DeclareMathOperator*{\argmax}{arg\,max}

\usepackage[nameinlink]{cleveref}

\crefname{assumption}{assumption}{assumptions}
\Crefname{assumption}{Assumption}{Assumptions}
\creflabelformat{equation}{#1#2#3}
\crefname{equation}{eq.}{eqs.}
\Crefname{equation}{Eq.}{Eqs.}
\Crefname{section}{\S}{\S}
\crefname{figure}{fig.}{figs.}
\Crefname{figure}{Fig.}{Figs.}
\crefname{algorithm}{alg.}{algs.}
\Crefname{algorithm}{Alg.}{Algs.}

\newtheorem{assumption}{Assumption}

\newcommand{\tv}{\mathrm{TV}}

\usepackage[acronym,smallcaps,nowarn]{glossaries}
\newacronym{ATE}{ate}{average treatment effect}
\newacronym{IPW}{ipw}{inverse propensity weighting}
\newacronym{TV}{tv}{total variation}
\newacronym{MLE}{mle}{maximum likelihood estimation}
\newacronym{DRO}{dro}{distributionally robust optimization}
\newacronym{RCI-NC}{rci-nc}{robust causal inference with noisy covariates}

\newcommand{\g}{\, | \,}
\newcommand{\s}{\, ; \,}
\newcommand{\dif}{\, d}

\newcommand{\vlambda}{\boldsymbol{\lambda}}

\newcommand{\vv}{\mathbf v}

\newcommand{\br}{\mathbb{R}}

\DeclareMathOperator{\logit}{logit}

\DeclareRobustCommand{\parhead}[1]{\textbf{#1}~}

%% file: sec/sec_abstract.tex

Causal inference from observational datasets often relies on measuring
and adjusting for covariates. In practice, measurements of the
covariates can often be noisy and/or biased, or only measurements of
their proxies may be available. Directly adjusting for these imperfect
measurements of the covariates can lead to biased causal estimates.
Moreover, without additional assumptions, the causal effects are not
point-identifiable due to the noise in these measurements. To this
end, we study the partial identification of causal effects given noisy
covariates, under a user-specified assumption on the noise level. The
key observation is that we can formulate the identification of the
average treatment effects (ATE) as a robust optimization problem. This
formulation leads to an efficient robust optimization algorithm that
bounds the ATE with noisy covariates. We show that this robust
optimization approach can extend a wide range of causal adjustment
methods to perform partial identification, including backdoor
adjustment, inverse propensity score weighting, double machine
learning, and front door adjustment. Across synthetic and real
datasets, we find that this approach provides ATE bounds with a higher
coverage probability than existing methods.


%% file: sec/sec_intro.tex
\section{Introduction}

Estimating the causal effect of an intervention is a problem that
arises in countless domains, with examples including identifying the
effect of medical treatments~\citep{connors1996effectiveness},
evaluating the effectiveness of recommender
systems~\citep{schnabel2016recommendations,wang2020causal}, and
assessing the impact of educational
methods~\citep{gustafsson2013causal}.  In many of these settings, the
challenge is to identify causal effects from observational data, and a
core problem is that naive inference can be biased by
\emph{confounders}, which are variables that affect both the
intervention and the outcomes. For example, in identifying the effect
of college education on earnings for students, the scholastic ability
is a confounder~\citep{card1999causal}---it can affect both whether
the student can be admitted to a college and how much he/she may earn
after graduation. As a result, the observed increase in earnings
associated with attending college is confounded by the effect of the
scholastic ability and thus cannot accurately represent the causal
effect of college education.

A common approach to addressing confounding bias is aiming to measure
all of the confounders and adjust for them~\citep{imbens2015causal}.
In practice, however, measurements of the confounders can often be
noisy or biased. Moreover, sometimes we only have access to proxies of
the confounders. For instance, in the example of college education and
earnings, confounders are often measured via surveys and thus are
typically biased and incomplete---participants may not be willing to
discuss aspects of their family backgrounds or reveal their access to
alternative educational or career options. Some confounders are
difficult to measure by definition---for example, students' innate
cognitive abilities---and in such cases we generally only have access
to proxies.

While noisy covariates (including both noisy measurements and
confounder proxies) are generally needed to perform causal inference,
directly adjusting for such covariates can lead to biased causal
estimates~\citep{fuller2009measurement}. Moreover, it is well known
that, without further assumptions on the causal model, the causal
effects are not point-identifiable given only noisy
covariates~\citep{carroll2006measurement,schennach2016recent,ogburn2013bias,lockwood2016matching}.
In other words, with access to only noisy covariates, it may be
impossible to pinpoint the causal effect of interest even with
infinite data. How then can noisy covariates inform causal inference?

In this paper, we leverage the noisy covariates to perform partial
identification of the causal effects. Given a user-specified
assumption on the noise level, we develop an algorithm for partial
identification using robust optimization. This approach capitalizes on
two observations: (1) the causal effects of interest are identifiable
given the (unobserved) true joint distribution of treatments,
outcomes, and all (noiseless) covariates; (2) the dataset with noisy
covariates places constraints on what this joint distribution can be.
These observations allow us to turn the task of partial identification
into a robust optimization problem. 

In more detail, we formulate partial identification as the following
robust optimization problem. We first consider an uncertainty set of
all possible underlying joint distribution of treatments, outcomes,
and the noiseless covariates subject to the constraints. Then we find
the maximum (or minimum) possible causal effects that a distribution
in this set can plausibly lead to. This approach propagates the
uncertainty in the true data distribution (due to covariate noise)
downstream to the uncertainty in the causal estimation, leading to
partial identification intervals of the causal effects.

Taking this optimization perspective on partial identification, we
develop an algorithm that efficiently solves the robust optimization
problem and computes the bounds on the causal effect of interest. This
algorithm is applicable to a wide variety of causal adjustment
methods, including backdoor adjustment, frontdoor adjustment, inverse
propensity score weighting (IPW), and double machine learning, which
we demonstrate. Across simulated and real datasets, we find that this
approach can produce tight bounds on causal effects that cover the
true average treatment effect.


\noindent\parhead{Contributions.} We propose a robust optimization approach
to partial identification given noisy covariates. The key idea is to
formulate partial identification with noisy covariates as a robust
optimization problem. We provide an efficient algorithm to solve this
robust optimization program, thereby obtaining upper and lower bounds
on the causal effects. We demonstrate the general applicability of
this approach by applying it to a variety of causal adjustment
methods. Finally, we demonstrate the effectiveness of the approach
across empirical studies with synthetic and real data.

%% file: sec/sec_related_work.tex
\paragraph{Related work. }This work draws on several threads of
research in measurement noise, proxy variables, and robust
optimization.

The first is on measurement noise and proxy variables in causal
inference. This subject has a long history in the
literature~\citep{wickens1972note,frost1979proxy}, where there have
been a variety of proposals for recovering causal effects either
heuristically or with additional model
assumptions~\citep{carroll2006measurement,schennach2016recent,ogburn2013bias,lockwood2016matching}.
Recent examples include \citet{louizos2017causal}, who use
variational autoencoders as a heuristic way to recover the latent
confounders; and \citet{kallus2018causal}, who use matrix
factorization to infer the confounders from the noisy covariates
assuming that the data-generating process follows a linear outcome
model. Given proxy variables of unmeasured confounders,
\citet{kuroki2014measurement} and \citet{miao2018identifying} propose
specific technical conditions under which causal effects can be
restored. These results have been extended to a variety of other
settings~\citep{tchetgen2020introduction,shi2020multiply,cui2020semiparametric,shpitser2021proximal,dukes2021proximal,ying2021proximal,shi2021proximal}.
Finally, \citet{imai2010causal} seek to partially identify ATE under
measurement error using constrained linear optimization. More
recently,
\citet{finkelstein2020partial,duarte2021automated,zhang2021partial,zhang2021non,balke1994counterfactual,balke1997bounds,ramsahai2012causal,bonet2013instrumentality,heckman2001instrumental,sachs2020symbolic,geiger2013quantifier}
develop optimization formulations for partial identification. Most of this work focuses on discrete variables under settings including unobserved confounding and/or measurement error.
Our work is complementary to this work; we focus on noisy
covariates but can
handle certain settings with continuous variables, without relying on
additional compliance assumptions.



Noisy covariates or proxy variables are not generally sufficient to
identify causal effects as they violate the ``no unobserved
confounders'' assumption. Therefore, handling noisy covariates relates
to sensitivity analyses that seek partial identification; i.e., bounds
on the average treatment effect (ATE)
\citep{liu2013introduction,richardson2014nonparametric,imbens2003sensitivity,veitch2020sense,dorie2016flexible,cinelli2020making,cinelli2019sensitivity,franks2019flexible,shen2011sensitivity,hsu2013calibrating,bonvini2020sensitivity,rosenbaum2010design,zhao2017sensitivity,yadlowsky2018bounds,zhang2020bounding,yin2021conformal}.
In this vein, the work of \citet{yadlowsky2018bounds} is 
most related. They propose a loss-minimization approach that
quantifies bounds on the conditional average treatment effect (CATE).
Their approach requires the unobserved confounder to satisfy a
constraint that bounds the effect on the odds of treatment selection.

The second thread of related work concerns robust optimization, which
is a core ingredient of the robust causal inference approach we
develop. We adopt a minimax formulation of a two-player game where the
uncertainty is adversarial, and one minimizes a worst-case objective
over a feasible set~\citep{BEN:09, bertsimas2011theory}.  For example,
the noise may be contained in a unit-norm ball around the input data.
To solve the robust optimization problem, we build on a recent line of
work on distributionally robust optimization (DRO) which assumes that
the uncertain distributions underlying the data have support within a
certain set~\citep{namkoong2016stochastic, duchi2018learning,
Li:2019}.

%% file: sec/sec_prelim.tex
\section{Preliminaries: Potential Outcomes, ATE Estimation, and Noisy
Covariates}

In this section, we set up the causal inference problem with noisy
covariates and formalize the assumptions required by the partial
identification of ATE.

\noindent\parhead{Potential outcome and ATE estimation. } Let $D^\ast = (X,
Y, Z)$ denote a dataset, where $X$ represents a vector of (possibly
unobserved) noiseless covariates. Denote $Z$ as a binary treatment
random variable, with $0$ and $1$ being the labels for control and
active treatments, respectively. Further, let $Y$ denote the outcome.
We use the potential outcomes notation to define causal quantities
\citep{neyman1923applications, rubin1974estimating}: For each
realization of the level of treatment, $z\in\{0,1\}$, we assume that
there exists a potential outcome, $Y(z)$, representing the outcome had
the subject been given treatment $z$ (possibly contrary to fact).
Then, the observed outcome is $Y=Y(Z)=ZY(1)+(1-Z)Y(0)$. We focus on
estimating the average treatment effect (ATE):
\begin{equation}\label{eq:ate}
     \tau = \E[Y(1) - Y(0)].
\end{equation}

To focus on the causal inference challenge due to noisy covariates, we
assume that the ATE is identifiable with the (potentially unobserved) noiseless data.
\begin{assumption}[Identifiability of ATE given noiseless
covariates]\label{assump-ignorable} The ATE is
identifiable~\citep{pearl1995causal} given the (unobserved) noiseless
covariates, in addition to the observed treatment and outcome, namely
the ATE can be written as a functional of the joint distribution of
$D^\ast = (X, Y, Z)$.
\end{assumption}

\Cref{assump-ignorable} ensures the identifiability of ATE if we had
access to the noiseless covariates. This assumption is satisfied when
the noiseless covariates $X$ meet identification conditions
for ATE such as the backdoor criterion, the
positivity condition and weak
unconfoundedness~\citep{rosenbaum1983central}, and
the frontdoor criterion~\citep{pearl2009causality}.

\noindent\parhead{Noisy covariates.} Though the noiseless data $D^\ast = (X,
Y, Z)$ can identify the ATE, we often do not have access to such a
dataset. Instead, we only have access to a dataset with noisy
covariates $\tilde{X}$. We denote the observed dataset as $D=
(\tilde{X}, Y, Z)$.

The noisy covariates $\tilde{X}$ shall potentially still provide
information about $X$ despite the noise. To describe to what extent
can $\tilde{X}$ inform $X$, we rely on an assumption about the noise
level.
\begin{assumption}[Noise level]\label{assump-TV} The TV distance
between the distributions of $X$ and $\tilde{X}$, conditional on the
treatment variable $Z$, is bounded by a constant~$\gamma_z$:
\begin{align}
\tv(p_{x|z}, p_{\tilde{x}|z}) \leq \gamma_z, z \in \{0,1\},
\label{eq:assump-TV}
\end{align}
where $p_{x|z}$ is the distribution of the unobserved noiseless
covariates for the treatment or control group, $X | Z = z \sim
p_{x|z}$, $p_{\tilde{x}|z}$ is the distribution of the noisy
covariates $\tilde X$, $\tilde{X}| Z=z \sim
p_{\tilde{x}|z}$, and the TV distance between two probability
distributions $p$ and $q$ is defined as follows:
\begin{align}
\label{eq:TV-def}
    \tv(p, q) &= \inf_{\pi} \E_{X,Y\sim \pi(x,y)} [\Ind(x \neq y)] \\
&\mathrm{s.t.} \int \pi(x, y)dy = p(x), \int \pi(x, y)dx = q(y),\nonumber
\end{align}
where $\pi$ represents a coupling between $p$ and
$q$~\citep[cf.][]{villani2008optimal}.
\end{assumption}

\Cref{assump-TV} provides a convenient way to characterize how far
away the noisy covariates $\tilde{X}$ are from their (unobserved)
noiseless counterpart $X$, where the further away the noisy covariates
$\tilde{X}$ is from $X$, the less informative $\tilde{X}$ is for
causal inference. The upper bound of the TV distance $\gamma_z$ in
\Cref{assump-TV} is a user-specified parameter, which is often
specified by domain experts or estimated from auxiliary datasets. In
particular, the TV distance as a distance to quantify the noise level
in \Cref{assump-TV} because of the computational tractability of TV
distance in robust optimization, which we will detail in
\Cref{sec:methods,app:dro-general}.

Many existing noise models imply the TV bound on distributions in
\Cref{assump-TV}. We illustrate a few of these models below.
\begin{enumerate}[leftmargin=*]
\item \textit{Huber contamination model.} Suppose the noisy covariates
deviate from the noiseless covariates following the Huber
contamination model \[p_{\tilde{x}|z} = (1-\gamma_z) p_{x|z} + \gamma_z
h_{\tilde{x}|z},\] where $h_{\tilde{x}|z}$ can be any arbitrary
distribution. Then the noisy covariates satisfy $\tv(p_{x|z},
p_{\tilde{x}|z})~\leq~\gamma_z.$
\item \textit{Misclassification error. } Suppose the noisy covariates
$\tilde{X}$ are discrete and their misclassification error satisfies
\[\max_x \left|P(X=x|Z=z) - P(\tilde{X}=x|Z=z)\right| <\gamma_z, \quad z\in \{0,1\}.\]
Then we also have $\tv(p_{x|z}, p_{\tilde{x}|z}) \leq \gamma_z.$
\item \textit{Exponential tilting model.} Suppose the distribution of
$\tilde{X}$ is an exponential tilted version of $X$, and both
distributions belong to the exponential family,
\[p(x|z) = \exp(\eta(\theta_z)\cdot T(x) + A(\theta_z) + B(x)),\qquad p(\tilde{x}|z) = \exp(\eta(\tilde{\theta}_z)\cdot T(\tilde{x}) + A(\tilde{\theta}_z) + B(\tilde{x})).\]
Then $\tv(p_{x|z}, p_{\tilde{x}|z}) \leq \gamma_z\triangleq 
\sqrt{\frac{1}{2} D_{A}(\tilde{\theta}_z,\theta_z)},$ where
$D_A(\cdot)$ is the Bregman divergence with function
 $A$~\citep{banerjee2005clustering}. 

\item \textit{Other models.} For general noise models for $\tilde{X}$,
one can approximately estimate the TV bound $\gamma_z$ in
\Cref{assump-TV} by drawing samples from $p_{\tilde{x}|z}$ and
$p_{x|z}$ respectively, calculating the KL divergence
estimates~\citep{perez2008kullback,belghazi2018mutual}, and applying Pinsker's inequality.
\end{enumerate}



Finally, \Cref{assump-TV} specifies a noise-level assumption on the
conditional distributions $p_{x|z}$ and $p_{\tilde{x}|z}$. One can
similarly specify the noise level for the marginal
distributions of the covariates, $p_{x}, p_{\tilde{x}}$, or the
conditional distributions given both the treatments and the outcomes,
$p_{x|y,z}, p_{\tilde{x}|y,z}$. Here we consider the use of  $p_{x|z}$ and $p_{\tilde{x}|z}$ as a demonstrative example, formulating
the partial identification task into a robust optimization problem in
\Cref{sec:methods}. Similar derivations apply to versions of
\Cref{assump-TV} with other distributions.

%% file: sec/sec_methods-main.tex

\section{Partial Identification with Noisy
Covariates}\label{sec:methods}

Though \Cref{assump-ignorable} ensures that the ATE is identifiable
given the noiseless covariates $X$, the ATE is not
point-identifiable without further assumptions given only the noisy
covariates $\tilde{X}$---there exist many
values of ATE that are all compatible with the observed distribution
of
$D$~\citep{carroll2006measurement,schennach2016recent,ogburn2013bias,lockwood2016matching}.

Given this lack of point identifiability, we focus on partial
identification of ATE. Instead of providing a point estimate of the
ATE, we aim to bound the ATE given the dataset with noisy covariates
$D= (\tilde{X}, Y, Z)$. In particular, we develop an optimization
approach to partial identification. The key idea is to cast the task
of partial identification as a robust optimization problem. We
consider the set of all joint distributions of the (unobserved)
noiseless data $P(X, Y, Z)$ that are compatible with the observed
noisy data $P(\tilde{X}, Y, Z)$ under the noise-level assumption
(\Cref{assump-TV}). Then the minimum and maximum value of ATE
resulting from these joint distributions shall bound the ATE. It turns
out that finding the minimum and maximum can be turned into a robust
optimization problem, for which we develop an algorithm to solve.



In the rest of this section, we begin with the parametric approach to
estimate ATE given noiseless data $D^\ast= (X, Y, Z)$; it can be
written as an optimization problem as the parameter model is fitted
via maximum likelihood. We then expand this optimization problem to
consider the dataset with noisy covariates~$D= (\tilde{X}, Y, Z)$,
which results in a robust optimization problem. 
We will
derive an efficient algorithm to solve the optimization and
demonstrate its general applicability to common causal adjustment
methods in \Cref{sec:examples}.

\subsection{The parametric modeling approach to ATE estimation}

\label{subsubsec:mle-ate}

We begin with estimating the ATE assuming oracle access to
noiseless data $D^\ast= (X, Y, Z)$.\footnote{The dataset contains $n$
i.i.d. data points $\{X_i, Y_i, Z_i\}_{i=1}^n$. We suppress the data
index for notation simplicity.} The ATE is identifiable given $D^\ast$
due to \Cref{assump-ignorable}. 

We adopt a parametric modeling approach to ATE estimation, where we
posit a parametric model for the joint distribution $p_{x,y,z}$ or its
components required by the identification formula. Specifically, we
first posit a parametric model for the joint distribution or its
relevant conditionals. For example, we may posit that the joint
distribution $p_{x,y,z}$ follows a parametric model
$\{p_{\theta}(x,y,z):\theta\in\Theta\}$, where $\Theta$ is the
parameter space. As another example, one may posit a parametric model
only for a conditional component of $p_{x,y,z}$, e.g. $p_\theta(x,y,z)
= p_{x} \times p_\theta(z|x)\times p_{y|x,z}$, where the conditional
$p_{z|x}$ follows a statistical model parameterized by $\theta$,
$\{p_{\theta}(z|x):\theta\in\Theta\}.$ Given the parametric model, we
find the likelihood maximizing parameter $\theta$
\begin{align}
\label{eq:max-ll}
\hat{\theta} = \argmax_\theta L_n(\theta\s p_{x,y,z}),
\end{align}
where $L_n(\theta\s p_{x,y,z})\triangleq \E_{p_{x,y,z}}[\log
p_\theta(x,y,z)]$ is the likelihood of the data $D^\ast=(X,Y,Z)$ at
parameter $\theta$. Finally, we plug in the fitted parametric model
for causal estimation
\begin{align}
\label{eq:ate-id}
\hat{\tau} = Q(p_{\hat{\theta}}(x,y,z)), 
\end{align}
where $p_{\hat{\theta}}(x,y,z)$ is the joint distribution of $(X, Y,
Z)$ implied by the posited statistical model at the optimal parameter
$\hat{\theta}$, and $Q(\cdot)$ is the causal identification functional
mapping the joint distribution of $(X, Y, Z)$ to the ATE $\tau$.

As an example, suppose we adopt the backdoor adjustment for
estimation. We first posit a parametric model for $p_{y|x, z}$ with
density $p_\theta(y|x, z) = \mathcal{N}(f(x, z; \theta), 1^2),$ find
the maximum likelihood parameters $\hat{\theta}$ by maximizing
$L_n(\theta; D^\ast)$, the Gaussian likelihood of the $n$ data points
in $D^\ast$ given parameter $\theta$, and finally calculate the ATE
estimate following the backdoor adjustment~$\hat{\tau} = \E_X[f(X, 1;
\hat{\theta})] -
\E_X[f(X, 0;\hat{\theta})]$.



\subsection{Partial identification as robust optimization}

The parametric approach to ATE estimation relies on having access to
noiseless covariates $X$. However, we often only have access to the
dataset with noisy covariates $D=(\tilde{X}, Y, Z)$, and the ATE is no
longer point identifiable; they may only partially identify the ATE.
Then how can we extend the parametric approach to partially identify
the ATE?

\noindent\parhead{Partial identification of ATE as a robust optimization.} To
perform partial identification, we extend the optimization problem of
\Cref{eq:max-ll} to a robust optimization. 
The key observation is that,
though the noiseless data distribution $p_{x,y,z}$ is unobserved, the
observed noisy data distribution $p_{\tilde{x}, y,z}$, together with
the noise-level assumption (\Cref{assump-TV}), characterizes an
uncertainty set of $p_{x,y,z}$, which further leads to an uncertainty
set of ATE, following the same identification formula in
\Cref{eq:ate-id}. If the uncertainty set of $p_{x,y,z}$ contains the
true $p_{x,y,z}$, then its resulting uncertainty set for the ATE shall
also contain the true ATE. In other words, the maximum and minimum of
this ATE uncertainty set  bound the true ATE, hence partial
identification.

Formally, we obtain the partial identification interval by solving the
following optimization problem analogous to the one in the parametric
approach (\Cref{eq:max-ll,eq:ate-id}). Denote the uncertainty set of
$p_{x,y,z}$ as $\mathcal{P}_{X,Y,Z}$. Then the lower bound of ATE
$\hat{\tau}_L$ is obtained by
\begin{align}
\hat{\tau}_L = \min_{p_{x,y,z}\in \mathcal{P}_{X,Y,Z}} &
Q(p_{\hat{\theta}}(x,y,z))\label{eq:ate-general-param}\\
\mathrm{s.t.} \quad & \hat{\theta} = \argmax L_n(\theta\s
p_{x,y,z}),\label{eq:MLE-grad-h}
\end{align}
which is a form of a distributionally robust optimization (DRO). We
can similarly obtain the upper bound $\hat{\tau}_U$ by replacing
$\min$ with $\max$, and the partial identification interval estimate
for the ATE $\tau$ is $[\hat{\tau}_L, \hat{\tau}_U]$. 
It is similar to
\Cref{eq:max-ll,eq:ate-id}: the parametric model is similarly placed
on the noiseless data $p_{x,y,z}$. The only difference is that
$p_{x,y,z}$ is unobserved; we have to calculate
\Cref{eq:max-ll,eq:ate-id} for all possible $p_{x,y,z}$ within the
uncertainty set $\mathcal{P}_{X,Y,Z}$. This formulation of partial identification as robust optimization
produces tight partial identification bounds. The tightness is
achieved by construction, as any $p_{x,y,z}$ that achieves the minimum
and maximum value of the objective is compatible with the observed
data and the posited statistical model due to the constraint of
$p_{x,y,z}\in \mathcal{P}_{X,Y,Z}$. Below we discuss some practical aspects of partial identification:
constructing the uncertainty set, solving the robust optimization
problem, and statistical inference of the partial identification
bounds.





\noindent\parhead{The uncertainty set $\mathcal{P}_{X,Y,Z}$.} To construct
the uncertainty set of $p_{x,y,z}$, we focus on characterizing
$p_{x|y,z}$, the conditional distribution of the noiseless covariates
$X$ given $Y,Z$. The reason is that the conditional distribution
$p_{x|y,z}$, along with treatment and outcome distribution $p_{y,z}$, fully determines the joint
$p_{x,y,z} = p_{x|y,z}\times p_{y,z}$. Thus the ATE can be identified
by \Cref{eq:ate-id} under \Cref{assump-ignorable}.

To construct the uncertainty set for $p_{x|y,z}$, we resort to
\Cref{assump-TV}, which requires that $p_{x|z} \in \{\bar{p}_{x|z}:
\tv(\bar{p}_{x|z}, p_{\tilde{x}|z}) \leq \gamma_z\}$ for $z \in
\{0,1\}$. Let us denote  $\boldsymbol{\gamma} = (\gamma_z)_{z\in\{0,1\}}$. Thus, by the chain rule, the uncertainty set of $p_{x,y,z}$
is
\begin{align*}
\mathcal{P}_{X,Y,Z}(p_{\tilde{x},y,z}\s\boldsymbol{\gamma}) =
\left\{p_{y,z}\times \bar{p}_{x|y,z}: \tv\left(\int
\bar{p}_{x|y,z}\times p_{y\mid z} dy, p_{\tilde{x}|z}\right)
\leq \gamma_z,  \forall y \in \mathcal{Y}, z \in \{0,1\}\right\}.
\end{align*}

\glsreset{DRO}

\sloppy
\noindent\parhead{Solving the robust optimization problem.}
\Cref{eq:ate-general-param,eq:MLE-grad-h} define a
distributionally robust optimization (DRO) problem, which generally takes the
form of a minimax optimization, $\min_{\theta \in \Theta} \max_{q:
D(q,p) \leq \gamma} \; \E_{X, Y \sim q}[l(\theta, X,Y)],$ where $D$ is
some divergence metric between the distributions $p$ and $q$, and
$l:\Theta \times \mathcal{X} \times
\mathcal{Y} \to \mathbb{R}$~\citep{duchi2018learning}. Such problems
can be solved by re-writing \Cref{eq:ate-general-param,eq:MLE-grad-h}
using via a Lagrangian formulation:
\begin{align}
\hat{\tau}_L =\min_\theta \min_{\substack{p_{x,y,z} \in\\
\mathcal{P}_{X,Y,Z}(p_{\tilde{x},y,z}\s\boldsymbol{\gamma})}}
\max_{\lambda\geq 0} \quad Q(p_{\theta}(x,y,z)) - \lambda \cdot
L_n(\theta\s p_{x,y,z}),
\end{align}
when the function $L_n(\cdot)$ is upper bounded. We can then apply
existing methods that efficiently and optimally solve the DRO problem
different divergence metrics
$D$~\citep{namkoong2016stochastic,Li:2019, Esfahani:2018}, equipped
with finite-sample convergence rates analyzed in
\citet{duchi2018learning}.  (See \Cref{app:dro-general} for details.)


\noindent\parhead{Statistical inference of the partial identification bounds.
} The robust optimization problem
(\Cref{eq:ate-general-param,eq:MLE-grad-h}) produces point estimates
for the partial identification bounds of ATE. To assess the sampling
uncertainty of these bounds, one can invoke standard statistical
inference tools~\citep{duchi2018learning}. Specifically, we consider a
separate TV ball around the observed data distribution
$\{\bar{p}_{\tilde{x},y,z}:
\tv(\bar{p}_{\tilde{x},y,z}, p_{\tilde{x},y,z})\leq \rho/n\}$, where
$n$ is the sample size and~$\rho = \chi^2_{1,1-\alpha}$ is the
$(1-\alpha)$-quantile of the $\chi^2_1$ distribution. We can then
obtain upper and lower confidence limits for $\hat{\tau}_L$ and
$\hat{\tau}_U$. For instance, for the lower bound of ATE
$\hat{\tau}_L$, its upper and lower confidence limits are
\begin{align*}
u_{\hat{\tau}_L} = \min_\theta \min_{\substack{p_{x,y,z} \in\\
\mathcal{P}_{X,Y,Z}(\bar{p}_{\tilde{x},y,z}\s\boldsymbol{\gamma})}}
\max_{\substack{\tv(\bar{p}_{\tilde{x},y,z}, p_{\tilde{x},y,z})\\\leq
\rho/n}}
\max_{\lambda\geq 0} \quad Q(p_{\hat{\theta}}(x,y,z)) - \lambda \cdot
L_n(\theta\s p_{x,y,z}),\\
l_{\hat{\tau}_L} = \min_\theta \min_{\substack{p_{x,y,z} \in\\
\mathcal{P}_{X,Y,Z}(\bar{p}_{\tilde{x},y,z}\s\boldsymbol{\gamma})}}
\min_{\substack{\tv(\bar{p}_{\tilde{x},y,z}, p_{\tilde{x},y,z})\\\leq
\rho/n}}
\max_{\lambda\geq 0} \quad Q(p_{\hat{\theta}}(x,y,z)) - \lambda \cdot
L_n(\theta\s p_{x,y,z}).
\end{align*}
Similarly, one can obtain the upper and low confidence limits of the
upper bound $\hat{\tau}_U$. Importantly, these confidence limits
$[l_{\hat{\tau}_L}, u_{\hat{\tau}_L}]$ quantify the sampling
uncertainty of $\hat{\tau}_L$ because we do not have access to the true
distribution $p_{\tilde{x},y,z}$. In contrast,
the partial identification bounds $[\hat{\tau}_L, \hat{\tau}_U]$
quantify the identification uncertainty of $\tau$ due to noisy
covariates. As the sample size $n$ increases, the confidence intervals
$[l_{\hat{\tau}_L}, u_{\hat{\tau}_L}]$ and $[l_{\hat{\tau}_U},
u_{\hat{\tau}_U}]$ shrink to a point mass, but the identification
interval $[\hat{\tau}_L, \hat{\tau}_U]$ does not shrink.


In more detail, suppose the propensity score given all observed
covariates is $e(X_{\mathrm{inc}}) \triangleq P(Z=1\g
X_{\mathrm{inc}})$, where $X_{\mathrm{inc}}$ denotes all the observed
covariates, which may not include all confounders and satisfy weak
unconfoundedness~\citep{imbens2015causal}. Further denote the
propensity score given all confounders $e(X_{\mathrm{full}})
\triangleq P(Z=1\g X_{\mathrm{full}})$, where $X_{\mathrm{full}}$
satisfy weak unconfoundedness $Z\perp Y(1), Y(0)\g X_{\mathrm{full}}$.
Then the robust optimization approach to partial identification can be
used to obtain ATE bounds under the sensitivity assumptions like
$\tv(p_{e(x_{\mathrm{inc}})}, p_{e(x_{\mathrm{full}})})\leq \gamma$
or~$\tv(p_{e(x_{\mathrm{inc}})|z}, p_{e(x_{\mathrm{full}})|z})\leq
\gamma_z$ for some constants $\gamma_z$, $\gamma$.

%% file: sec/sec_examples.tex
\section{Applications to Common Causal Adjustment Methods}\label{sec:examples}

In this section, we apply the general robust optimization strategy to
a variety of popular causal adjustment methods. In particular, we
instantiate the estimator
(\Cref{eq:ate-general-param}) for three adjustment methods, 
backdoor adjustment, inverse propensity weighting (IPW), and frontdoor
adjustment. (We further demonstrate the application to double machine
learning in \Cref{app:double-ml}.) For each adjustment method, we
first provide a brief review of the standard procedure, then
demonstrate how it can be augmented to perform partial identification given noisy covariates. Specifically, we write the objective
$Q(\cdot)$ and the likelihood constraint $L_n(\cdot)$ as a
functional of the unobserved conditional $p_{x|y,z}$, because the
uncertainty set of the joint $p_{x,y,z}$ is expressed in terms of
$p_{x|y,z}$. These steps will enable us to solve the robust
optimization problem by searching $p_{x|y,z}$ over the uncertainty
set.



\noindent\parhead{Backdoor adjustment.} Under the backdoor criterion, backdoor
adjustment estimates the potential outcomes $\E[Y(z)], z \in \{0,1\}$
by $\E[Y(z)]= \int\E[Y| Z=z, X=x]P(X=x) dx.$ If we had access to
the noiseless covariates $X$, we estimate the ATE by positing a
parametric model $\E[Y| Z=z, X=x] = g(x, z; \theta)$ and calculating
$\widehat{\tau} = \E[g(X, Z=1; \theta)-g( X, Z=0;\theta)]$.

Next we move from noiseless covariates to noisy ones. Given each
feasible $p_{x|y,z}$, we inherit the identification formula as the
optimization objective $Q(p_{\hat{\theta}}(x,y,z)) =
\E_{p_{x|z=1}}[g(X, Z=1;
\theta)]-\E_{p_{x|z=0}}[g( X, Z=0;\theta)]$, where $p_{x|z} = \int
p_{x|y,z} \times p_{y|z} \dif y$. Then the constraint is that $\theta$
maximizes the expected log-likelihood given the dataset:
$L_n(\theta\s p_{x,y,z}) = \E_{Y, Z}[\E_{p_{x|y,z}}[ \ell (f(X, Z;
\theta), Y)]]$.

\noindent\parhead{Inverse propensity weighting (IPW). }The application to the
IPW method share a similar spirit as backdoor adjustment except that
IPW estimates the potential outcome $Y(z), z\in \{0,1\}$ with a
different estimator $\E[Y(z)] = \E[ \frac{YZ}{P(Z=z|X)}]$. We estimate the ATE by positing a
parametric model on the propensity score, $Z| X \sim
\text{Bern}(f(\theta, X))$. Thus for each feasible $p_{x|y,z}$, we can
estimate the ATE by $Q(p_{\hat{\theta}}(x,y,z)) =\E_{p_{y,z}}
\E_{p_{x|y,z}}[\frac{YZ}{f(X;\theta)} - \frac{Y(1-Z)}{1-f(X;
\theta)}]$, which is also the objective of the robust optimization
problem. Then the constraint of this problem is that $\theta$
maximizes the likelihood of $p_{z|x}$, i.e. $L_n(\theta\s
p_{x,y,z})=\E_{z}[\E_{p_{x|z}}[ \mathrm{Bern}(Z\s f(X; \theta))]]$
with $p_{x|z} = \int p_{x|y,z}\times p_{y|z}\dif y.$

\noindent\parhead{Frontdoor adjustment. }Frontdoor adjustment is different
from the backdoor adjustment and IPW in that the covariates $X$ serve
as mediators between the treatment $Z$ and outcome $Y$. Frontdoor
adjustment gives the following estimator for potential outcomes,
$\E[Y(z)] = \E_{X\sim P(X| Z=z)}[\sum_{z'=0,1} \E[Y|X, Z=z']
P(Z=z')].$ Similar to backdoor adjustment, we can parameterize
$\E[Y|X=x, Z=z]=g(x,z\s\theta).$ Thus the ATE identification
functional is $Q(p_{\hat{\theta}}(x,y,z)) = \E_{X\sim
P(X|Z=1)}[\sum_{z'=0,1} g(X,z'; \theta) P(Z=z')] -  \E_{X\sim
P(X|Z=0)}[\sum_{z'=0,1} g(X,z'; \theta) P(Z=z')].$ As we use the same
parametric model as in backdoor adjustment, the constraint of the
robust optimization problem of the frontdoor adjustment is the same as
that of the backdoor adjustment, where $L_n(\theta\s p_{x,y,z}) =
\E_{Y, Z}[\E_{p_{x|y,z}}[ \ell (f(X, Z; \theta), Y)]]$.

%% file: sec/sec_experiment.tex
\vspace{-2mm}
\section{Experiments}\label{sec:experiment}

\begin{figure*}[!ht]
  \begin{center}
    \begin{tabular}{ccc}
           \includegraphics[width=0.27\textwidth]{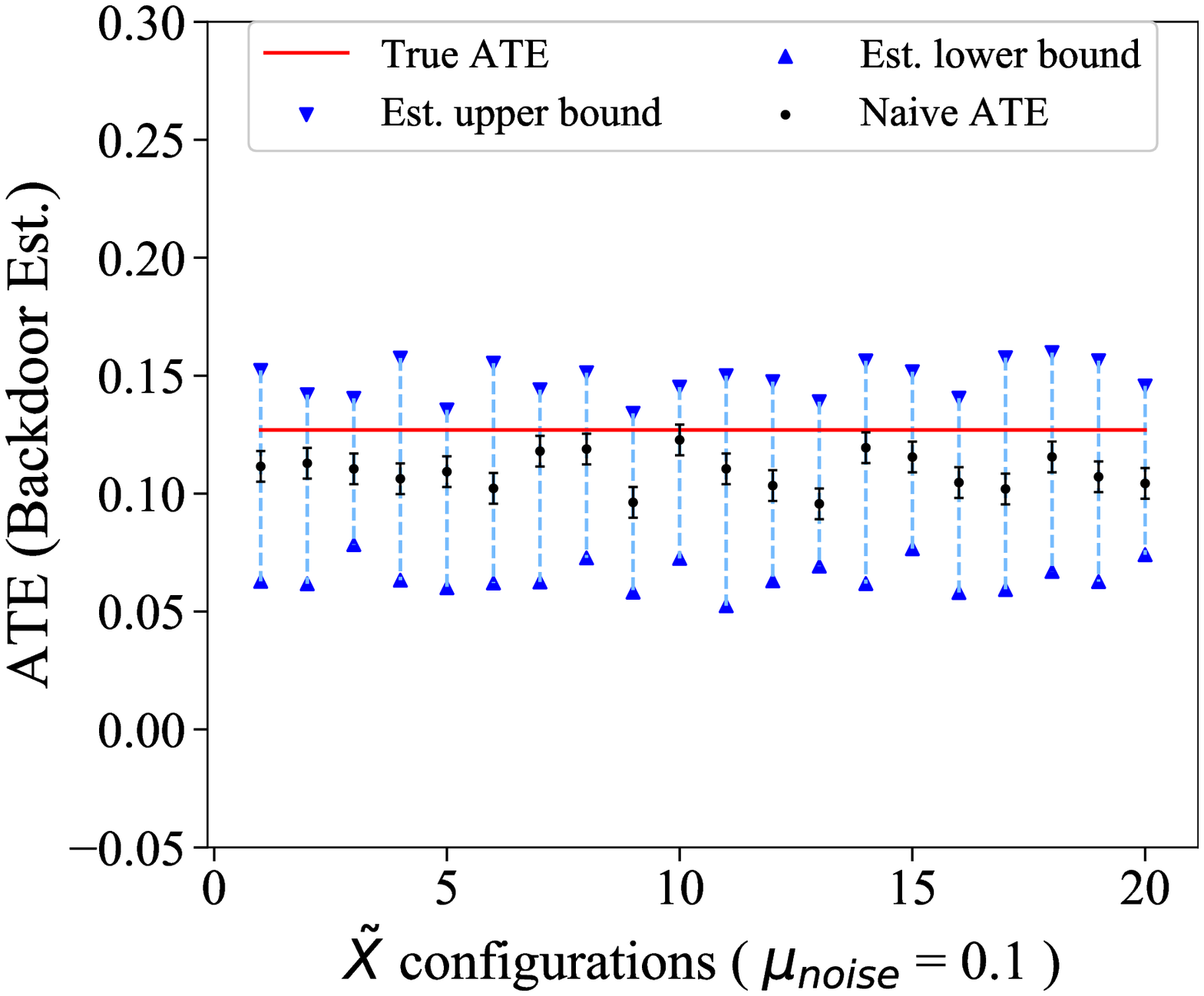} & \includegraphics[width=0.27\textwidth]{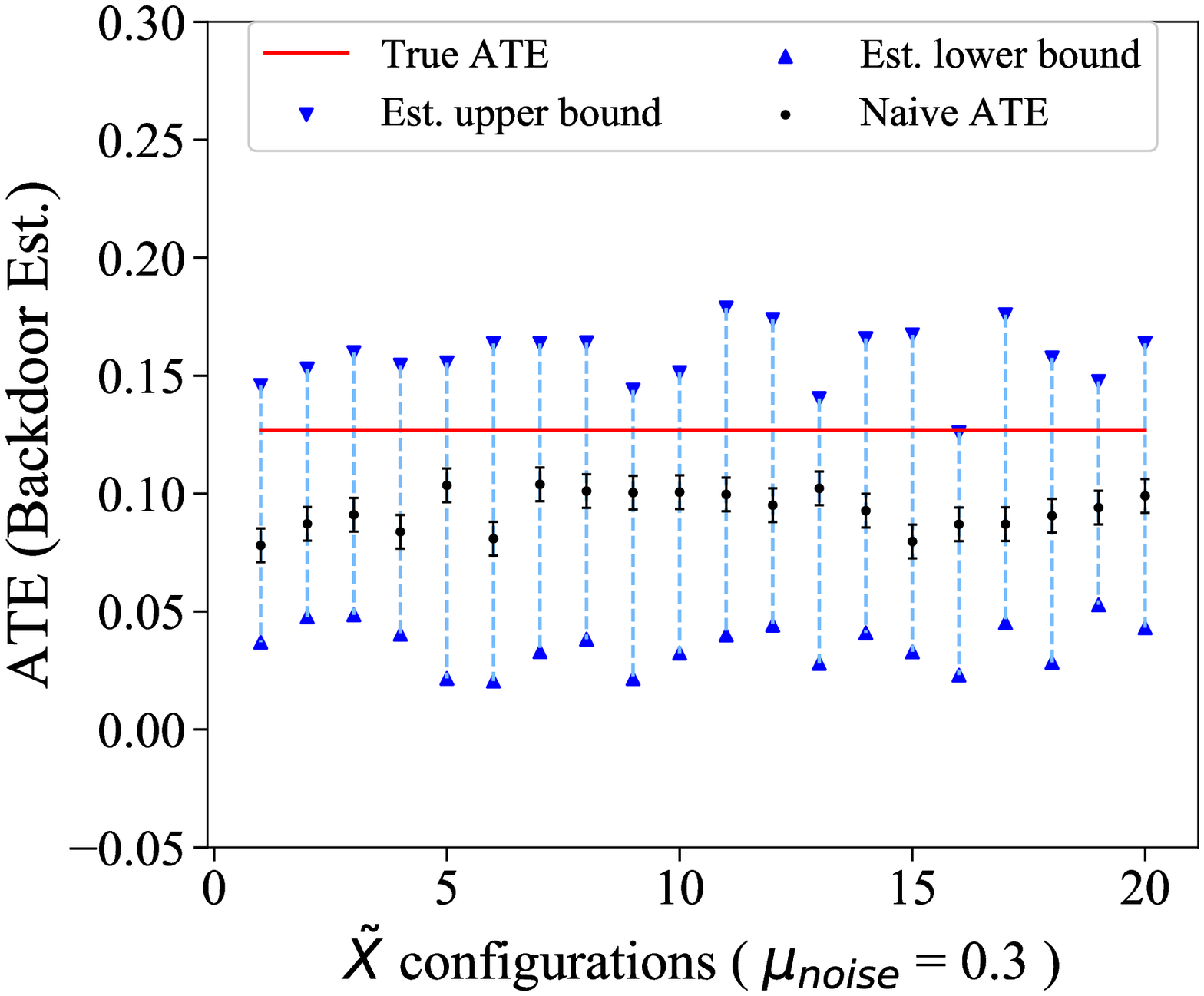} & \includegraphics[width=0.27\textwidth]{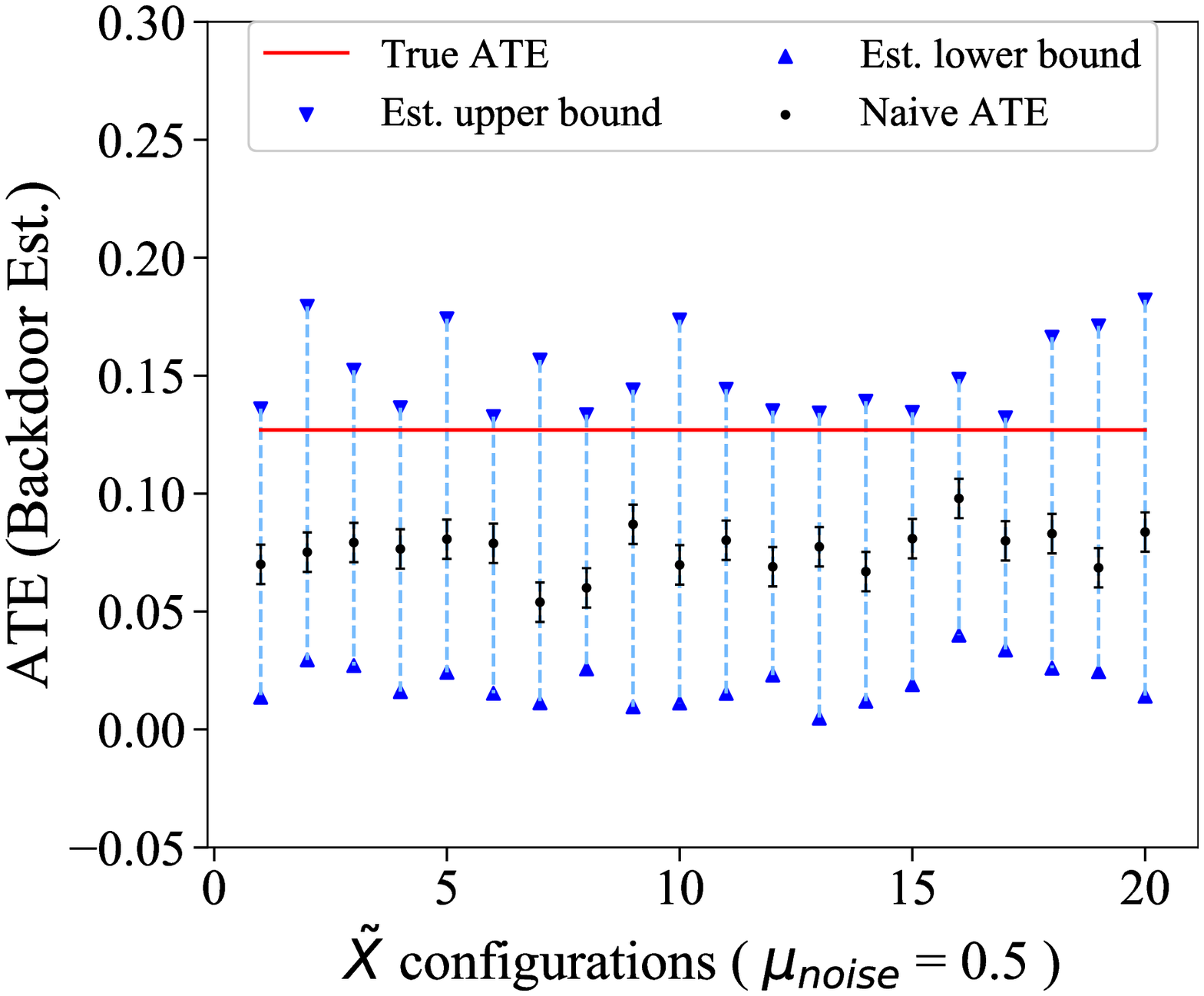} \\
          \; (a) & \;\; (b) & \;\;\; (c) 
    \end{tabular}
    \caption{Partial identification of ATE with backdoor adjustment estimators on synthetic dataset with binary outcome. The three noise levels have random Gaussian noise with mean = 0.1/0.3/0.5 and standard deviation = 0.5/0.5/1. In all plots, we compare RCI (this work) to the naive approach. The error bars indicate 95\% confidence interval of the naive ATE estimation over twenty trials. \textit{Intervals covering the true ATE is better.}\vspace{-8mm}} 
    \label{fig:backdoor-frontdoor-sim}
  \end{center}
\end{figure*}

We empirically evaluate the performance of the partial identification for noisy covariates via robust optimization (abbreviated as RCI) on a variety of simulated and real datasets. For each dataset, we synthetically generate noisy versions of it with different noise levels. For each noise level, we compute the noise strength $\gamma_z$ as the TV distance in \Cref{eq:assump-TV}, which is a parameter of RCI to estimate ATE. We study the performance of RCI applied to a variety of standard causal estimators, including the backdoor adjustment estimator, the IPW estimator and the frondoor adjustment, comparing them with a naive approach that employs the corresponding estimator directly applied to the noisy data. We find that RCI provides partial identification intervals with improved coverage properties than existing approaches, including the Causal Effect Variational
Autoencoder (CEVAE)~\citep{louizos2017causal}, while being not overly conservative (e.g. \Cref{fig:backdoor-frontdoor-sim}). We provide the details of the datasets, evaluation procedures and results in sequel. Further data and training details are in Appendix~\ref{app:exp}. 

\subsection{Synthetic data}


\noindent\parhead{Backdoor adjustment and IPW.} To evaluate the performance of the robust approach with the backdoor adjustment and IPW estimators, we synthetically generate two datasets with $X$ as the confounder. To demonstrate the variability of the estimated ATE intervals, we first synthetically generate a dataset with binary outcomes according to a logistic model. We further consider another synthetic dataset with a more complicated nonlinear outcome model and continuous outcomes, using the Kang and Schafer example \citep{kang2007demystifying}, which consists of four unobserved covariates $U_i \stackrel{iid}{\sim} N(0, I_4)$, $ i= 1,..., n$. The full data generation details are included in Appendix~\ref{app:synthetic-data-generation}.

\begin{table}[!ht]
\caption{Coverage probabilities for the partial identification interval via robust optimization with the backdoor adjustment and IPW estimators, and the naive approach and CEVAE (averaged over 100 trials, with standard error). (\textit{higher} is better).}
\label{table:sim-backdoor-ipw}
\begin{center}
\begin{small}
\begin{tabular}{l|c|c|cc}
\hline &
& & \multicolumn{2}{c}{RCI} \\
Noise & Naive & CEVAE & Backdoor Adj. &  IPW  \\
\hline
level 1 & 0.10 $\pm$ 0.09 & 0.46 $\pm$ 0.17 & \textbf{1.00 $\pm$ 0.00}  & \textbf{0.95 $\pm$ 0.02}  \\

level 2 & 0.00 $\pm$ 0.00 & 0.41 $\pm$ 0.05& \textbf{0.98 $\pm$ 0.01}  & \textbf{0.93 $\pm$ 0.03} \\

level 3 & 0.00 $\pm$ 0.00 & 0.42 $\pm$ 0.09 &  \textbf{0.97 $\pm$ 0.02} &  \textbf{0.91 $\pm$ 0.03} \\

level 4 & 0.00 $\pm$ 0.00 & 0.33 $\pm$ 0.03 &  \textbf{0.95 $\pm$ 0.02}  &  \textbf{0.87 $\pm$ 0.03} \\

level 5 & 0.00 $\pm$ 0.00 & 0.32 $\pm$ 0.14 &  \textbf{0.93 $\pm$ 0.03} &  \textbf{0.85 $\pm$ 0.04} \\
\hline
\end{tabular}
\end{small}
\end{center}
\vskip -5pt
\end{table}

\noindent\parhead{Frontdoor adjustment.} For front adjustment, we synthetically generate two datasets, with $X$ as the mediators. First, we generate a dataset with binary outcomes using a logistic model with a single mediator. We then generate another more complicated using a similar data generation as in \citet{jung2020estimating}. This data generation is more complicated with multiple mediators. 

\noindent\parhead{Noisy data generation.} Given the true covariates, we generate noisy covariates by synthetically adding a small amount of random noise. The ground truth covariates enables us to estimate the true ATE. For each selected example, we perturb it by adding a noise drawn from a Gaussian distribution to each dimension. We then evaluate the performance of the different algorithms ranging from small to large amounts of noise. The full generation details are in~\ref{app:synthetic-data-generation}.

\noindent\parhead{Evaluation and results.} To demonstrate the variability of the estimated ATE intervals, we plot the ATE intervals obtained by RCI and the true ATE. As a comparison, we also show the results of the naive ATE estimator, which estimates ATE directly using the noisy examples. The true ATE is calculated using the corresponding adjustment method and the noiseless covariates. We generate 2000 samples for each adjustment method, and generate 20 configurations of the noisy covariates for each noise level.

Figure~\ref{fig:backdoor-frontdoor-sim} shows the performance of different algorithms using the backdoor adjustment method. The results with frontdoor adjustment method are similar and included in Appendix~\ref{app-exp-results}. We observe that under the two outcome models, RCI can provide ATE intervals with a high coverage on the true ATE. However, the naive approach is very sensitive to the noise even for the low noise levels, and gives estimations that deviate from the true ATE as the noise level increases. Moreover, the RCI intervals are not overly conservative or trivial: they cover the true ATE without much overshooting.
\begin{table}[!t]
\caption{Coverage probabilities for the robust optimization approach with frontdoor adjustment, and the naive approach. (a) shows results with simulation data contains multiple mediators (mean and standard errors are averaged over 100 trials.). (b) shows results with the IHDP dataset (mean and standard errors are averaged over 50 trials.) (\textit{higher} is better).} \label{table:sim-frontdoor}
\begin{center}
\begin{small}
\begin{subtable}{.45\linewidth}
\centering
\begin{tabular}{l|cc}
\hline 
Noise  & Naive & RCI (Frontdoor)\\
\hline
level 1 &0.30 $\pm$ 0.15&\textbf{0.92 $\pm$ 0.03} \\

level 2 & 0.10 $\pm$ 0.09 & \textbf{0.87 $\pm$ 0.03} \\

level 3 & 0.00 $\pm$ 0.00 & \textbf{0.84 $\pm$ 0.04} \\

level 4 & 0.00 $\pm$ 0.00 & \textbf{0.82 $\pm$ 0.04}\\

level 5 & 0.00 $\pm$ 0.00 &  \textbf{0.81 $\pm$ 0.04} \\
\hline
\end{tabular}
\end{subtable}
\quad \quad
\begin{subtable}{.45\linewidth}
\centering
\begin{tabular}{l|cc}
\hline 
Noise  & Naive & RCI (Frontdoor)\\
\hline
level 1 &0.30 $\pm$ 0.15&\textbf{0.98 $\pm$ 0.02} \\

level 2 & 0.10 $\pm$ 0.09 & \textbf{0.96 $\pm$ 0.03} \\

level 3 & 0.00 $\pm$ 0.00 & \textbf{0.94 $\pm$ 0.04} \\

level 4 & 0.00 $\pm$ 0.00 & \textbf{0.94 $\pm$ 0.04}\\

level 5 & 0.00 $\pm$ 0.00 &  \textbf{0.92 $\pm$ 0.04} \\
\hline
\end{tabular}
\end{subtable}
\end{small}
\end{center}
\end{table}
We compute the true ATE coverage probability with the more complicated Kang and Schafer example (used for backdoor adjustment and IPW), and the second simulated dataset with multiple mediators for the frontdoor adjustment method. We also compare with CEVAE \citep{louizos2017causal}, which identifies ATE via back-door adjustment and models the noised covariates  as the proxy variables. A success cover means that the true ATE is contained in the estimated ATE interval by RCI, or by the 95\% confidence intervals of naive ATE. For CEVAE, at a specific noise level, we collect its ATE estimates over multiple datasets with noisy covariates. A success cover means the true ATE is within the range of estimates from the noisy datasets. We generate ten random noiseless datasets of true covariates with size 2000. For each noiseless dataset, we further generate ten equal-sized datasets with noisy covariates by drawing fresh noise samples. Therefore, the coverage probabilities are calculated over 100 pairs of true and noisy datasets. Table~\ref{table:sim-backdoor-ipw} and Table~\ref{table:sim-frontdoor} (\textit{left}) show the coverage probabilities using the three adjustment methods. We see that RCI is able to maintain a much higher coverage probability as the noise level increases.






\subsection{Real data case studies}

We further test the robust approach RCI on two case studies with real covariates, including an ACIC dataset and an IHDP dataset. Both datasets have been used for benchmarking various causal inference algorithms~\citep{shalit2017estimating,shi2019adapting,gupta2020estimating}.


\noindent\parhead{Case study 1: ACIC dataset.} We first use a dataset constructed for the Atlantic Causal Inference Conference (ACIC) 2019 Data Challenge based on the ``spambase'' dataset for spam email detection from UCI \citep{acic2019,DuaUCI}. This dataset consists of emails with an outcome of interest $Y$ being whether or not the email was marked as spam by a user. The treatment $Z$ represents whether or not the email contains more than a given threshold of capital letters, where this threshold is computed by a mean over the original dataset. There are 22 continuous covariates $X$ which are word frequencies given as percentages between 0 and 100. We generate our dataset directly using ACIC's data generating process, with a size of 2000 examples.  Given the true covariates, we further generate noisy covariates by synthetically adding a small amount of noise at random, using a similar procedure as for the synthetic data. Specifically, we generate five levels of Gaussian noise with mean = 0.1/0.2/0.3/0.4/0.5 and standard deviations at 0.5/0.5/1/1/1.


\begin{table}[!t]
\caption{Coverage probabilities for the robust optimization approach with the backdoor adjustment and IPW estimators, and the naive approach, using the ACIC dataset. (The results are averaged over 50 trials). (\textit{higher} is better).}
\label{table:acic-backdoor-ipw}
\begin{center}
\begin{small}
\begin{tabular}{l|c|c|cc}
\hline & &
& \multicolumn{2}{c}{RCI} \\
Noise & Naive & CEVAE & Backdoor Adj. &  IPW \\
\hline
level 1 & 0.02 $\pm$ 0.02& 0.81 $\pm$ 0.07& \textbf{1.00 $\pm$ 0.00}  & \textbf{1.00 $\pm$ 0.00} \\

level 2 & 0.00 $\pm$ 0.00& 0.73 $\pm$ 0.05& \textbf{0.98 $\pm$ 0.02}    & \textbf{0.98 $\pm$ 0.02} \\

level 3 & 0.00 $\pm$ 0.00 & 0.75 $\pm$ 0.10  & \textbf{0.94 $\pm$ 0.03} &  \textbf{0.92 $\pm$ 0.04}\\

level 4 & 0.00 $\pm$ 0.00 & 0.64 $\pm$ 0.02 & \textbf{0.94 $\pm$ 0.03}  &  \textbf{0.90 $\pm$ 0.04}\\

level 5 & 0.00 $\pm$ 0.00 & 0.64 $\pm$ 0.03 & \textbf{0.90 $\pm$ 0.04} &  \textbf{0.90 $\pm$ 0.04}\\
\hline
\end{tabular}
\end{small}
\end{center}
\vskip -7pt
\end{table}

\noindent\parhead{Case study 2: IHDP dataset.} For a second case study, we use a benchmark dataset introduced by~\citet{hill2011bayesian}, which is constructed from data obtained from the Infant Health and Development Program (IHDP). This dataset is based on a randomized experiment to measure the effect of home visits from a specialist on future test scores of children. The confounders $U$ correspond to collected measurements of the children and their mothers used during a randomized experiment that studied the effect of home visits by specialists on future cognitive test scores. We use samples from the NPCI package \citep{Dorie2016}, which converted the randomized data to an observational study by removing a biased subset of the treated group. The final dataset contains $747$ samples with $25$ covariates. We then simulate the mediator and the outcome using a procedure similar to \citet{hill2011bayesian, gupta2020estimating}. We generated the noisy covariates using the same five noise levels as the ACIC dataset. The full generation details are in~\ref{app:real-data-details}.

\noindent\parhead{Evaluation results.} We evaluated the naive approach, the RCI approach with the backdoor adjustment and IPW adjustment methods on the ACIC dataset. We also evaluated the naive approach and RCI with the frontdoor adjustment method on the IHDP dataset. Table~\ref{table:acic-backdoor-ipw} and Table~\ref{table:sim-frontdoor}(\textit{right}) show the coverage probabilities using these three adjustment methods. For both case studies, RCI is able to maintain a much higher coverage probability as the noise level increases, while the naive approach's estimates turn out to be very sensitive to the noise and have low coverage probabilities. Frontdoor adjustment method is able to achieve a higher coverage probability comparing to the synthetic data. This could be due to the fact that, in this data generation model, the outcome is linearly correlated with the mediator. As we also used a linear parameterized model, there is no model specification.

%% file: sec/sec_conclu.tex
\section{Conclusion} 
\label{sec:conclu}

This paper develops an approach to partial identification for noisy covariates via robust optimization. We show that partial identification can be formulated as a robust
optimization problem, which enables bounds on causal effects for parametric causal models. We then derive a variant of the projected gradient algorithm to  efficiently solve the robust optimization problem and
compute partial identification bounds on the causal effect of interest. We illustrate the wide applicability of our approach 
on a variety of causal adjustment methods, including the backdoor adjustment, inverse propensity weighting and the frontdoor adjustment. Numerical results across synthetic and real-world data show that this approach can effectively compute bounds for ATE with higher coverage than previous methods without being overly conservative.

%% file: sec/app_dro.tex
\section{Further Details about Solving the General DRO problem} \label{app:dro-general}

In this section,  we describe the details on solving the general DRO
problem \Cref{eq:ate-general-param} with \gls*{TV} distance using the
empirical Lagrangian formulation. 

Many existing works on DRO study how to solve the DRO problem for
different divergence metrics $D$. The robust optimization problem
(\Cref{eq:ate-general-param,eq:MLE-grad-h}) can be written in the form
of a DRO problem with TV distance by Lagrangian formulation.
\citet{namkoong2016stochastic} provide methods for efficiently and
optimally solving the DRO problem for $f$-divergences, and other work
has provided methods for solving the DRO problem for Wasserstein
distances~\citep{Li:2019, Esfahani:2018}. \citet{duchi2018learning}
further provide finite-sample convergence rates for the empirical
version of the DRO problem.

Below we describe the empirical Lagrangian formulation and adopt a
projected gradient-based algorithm to solve it and provide the
pseudo-code of the algorithm.

\subsection{Empirical Lagrangian formulation}

In a general form, we consider the parameterized ATE estimator $\hat
\tau$ as a general function of $\bar p_{x|y,z}, z\in \{0,1\}$,
$f_0(\bar p_{x|y,z=0},\bar p_{x|y,z=1};\theta)$, where $\theta$
denotes the parameters, and $f_0$ denotes the identification
functional of ATE, e.g. the $g$-formula. We write $f_0$ as a
functional of $p_{x|y,z}$ because, within the full joint $p_{x,y,z}$
that can enable ATE identification, $p_{x|y,z}$ is the only component
that is unobserved.

For simplicity of exposition, we first consider a special case of the
constraint in \Cref{eq:MLE-grad-h}, i.e. when we assume independent
Gaussian noise in the statistical model of $p_{x,y,z}$, e.g.
$\bar{p}_\theta(x|y,z) = \mathcal{N}(h_{\theta_1}(y,z), \theta_2^2)$.
In this case, the constraint in \Cref{eq:MLE-grad-h} becomes
equivalent to minimizing the mean squared error (MSE). Denote the MSE
as $f_1(\bar{p}_{x|y,z}; \theta)$. We then rewrite the constraint as
$f_1(\bar{p}_{x|y,z}; \theta) \leq \epsilon$, where $\epsilon>0$ is a
slack variable taking on a small positive value.

Then \Cref{eq:ate-general-param,eq:MLE-grad-h} becomes
\begin{align}\label{opt:general-inequalities}
    \begin{split}
        \min_{\substack{\theta, \bar{p}_{x|y,z} \in
        \mathcal{P}_{X|Y,Z}\\z=0,1}} \quad  &f_0(\bar p_{x|y,z=0},\bar
        p_{x|y,z=1};\theta) \\
        \quad \quad\textrm{  s.t.} \quad \quad\quad  & f_1(\bar
        p_{x|y,z=0},\bar p_{x|y,z=1};\theta)\leq \epsilon
    \end{split}
\end{align}

For simplicity, let $v(\bar p_{x|y,z=0}; \theta) = f_1-\epsilon$. Then
the Lagrangian of \Cref{opt:general-inequalities} is:
\begin{align*} L(\bar p_{x|y,z=0},\bar p_{x|y,z=1},
    \lambda;\theta)=f_0(\bar p_{x|y,z=0},\bar p_{x|y,z=1};\theta) +
    \left \langle \lambda, v(\bar p_{x|y,z=0},\bar p_{x|y,z=1};\theta)
    \right \rangle,
\end{align*} where $\lambda \geq 0$ is the Lagrange multiplier. Thus
the optimization problem of  \Cref{opt:general-inequalities} can be
rewritten as
\begin{align}\label{opt:general-dro-dual}
     \min_{\theta \in \Theta}\min_{\substack{\bar{p}_{x|y,z} \in
     \mathcal{P}_{X|Y,Z}\\z=0,1}} \max_{\lambda\geq 0} \quad L(\bar
     p_{x|y,z=0},\bar p_{x|y,z=1}, \lambda;\theta).
\end{align}

It remains to solve \Cref{opt:general-dro-dual}, for which we resort
to the empirical formulation of the DRO problem. Specifically, we
replace all expectations with expectations over empirical
distributions given a dataset of $n$ samples, i.e. $D = \{(\tilde X_1,
Y_1, Z_1), \ldots,(\tilde X_n, Y_n, Z_n)\}$. Specifically, we consider
the TV constraint between the respective empirical distributions of
$\bar{p}_{x|z}$ and $p_{\tilde{x}|z}$, as opposed to their population
version to which we do not have access. Such a TV constraint between
empirical distributions reduces to an $\ell_1$ norm constraint due to
the definition of the TV distance. This $\ell_1$ reduction is
particular suitable for efficient solving the DRO problem, which we
detail in \Cref{subsec:gda}.

In more detail, for each $z \in \{0,1\}$, let $n_z$ be the number of
samples with $Z_i=z$. Then we consider the empirical version of
$p_{\tilde{x}|z}\in \br^{n_z\times |\widetilde{\mathcal{X}}|}$ be a
probability table with $n_z$ rows and $|\widetilde{\mathcal{X}}_z|$
columns, where $\widetilde{\mathcal{X}}_z$ is set of (unique) values
taken by $(\tilde X_i)_{Z_i=z}$; its $(i,j)$ cell takes the value
$p_{\tilde{x}|z}^i = \frac{1}{n_z}$ if the $i$-th example satisfies
$Z_i = z$, $X_i$ takes the $j$th value in the
$\widetilde{\mathcal{X}}_z$ set. We then consider the empirical
distribution of $\bar{p}_{x|z}\in \br^{n_z}$ in a similar way and
rewrite the \gls*{TV} distance constraint as $\ell_1$ norm
constraints: $||\bar{p}_{x|z},p_{\tilde{x}|z}||_1 \leq 2 \gamma_z$ for
all $z \in \{0,1\}$.


Replacing all expectations with expectations over the appropriate
empirical distributions, we rewrite the constraints as $\ell_1$ norm
constraints on the empirical distribution of $X$ given $Z$. Then
\Cref{opt:general-dro-dual} is equivalent to:

\begin{align}\label{opt:general-dro-emp}
    \begin{split}
        \min_{\theta} \max_{\lambda \geq
        0}\max_{\substack{\bar{p}_{x|y,z},\\ z =0,1}} &L(\bar p_{x|y,z=0},\bar
        p_{x|y,z=1}, \lambda;\theta) \\ \text{s.t. } & ||\bar{p}_{x|z} -
        p_{\tilde{x}|z}||_1 \leq 2\gamma_z,\;\; || \bar{p}_{x|z} ||_1
        = 1, \quad \forall z \in \{0,1\}.
    \end{split}
\end{align}

\subsection{Projected GDA algorithm}

\label{subsec:gda}

To solve \Cref{opt:general-dro-emp}, we use a projected gradient descent ascent (GDA) algorithm, which is a simplified version of the algorithm introduced by \citet{namkoong2016stochastic} for solving general classes of DRO problems. Note that projections onto an $\ell_1$-ball can be done efficiently~\citep{duchi2008efficient}. We provide the pseudocode in \Cref{alg:general-dro-gda}. The implementation code will be made public.

\begin{algorithm}[H] 
\caption{Project GDA Algorithm for the general DRO formulation}
\label{alg:general-dro-gda}
\begin{algorithmic}[1]
\REQUIRE learning rates $\eta_\theta > 0$, $\eta_\lambda > 0$,  $\eta_z > 0, z=0,1$; upper bounds $\gamma_z>0, z=0,1$.
\vskip 0.05in
\vskip 0.05in
\FOR{$t = 1, \ldots, T$}
\vskip 0.05in
\STATE \textit{Descent step on $\theta$:} \\
 Compute $\theta^{(t+1)} \gets \theta^{(t)} - \eta_\theta\cdot \nabla_\theta L\left(\bar p_{x|y,z=0}^{(t)},\bar p^{(t)}_{x|y,z=1}, \lambda^{(t)};\theta^{(t)}\right)$
\vskip 0.05in
\STATE \textit{Ascent step on $\vlambda$}:\\

Compute $\vlambda^{(t+1)} \gets \vlambda^{(t)} + \eta_\theta \cdot \vv\left(\bar p_{x|y,z=0}^{(t)},\bar p_{x|y,z=1}^{(t)};\theta^{(t)}\right)$\\
\vskip 0.05in
\FOR{$z \in \{0,1\}$}
        \STATE \textit{Ascent step on} $\bar{p}_{x|z}$: Compute $\bar{p}_{x|z}^{(t+1)} \gets \bar{p}_{x|z}^{(t)} + \eta_z \cdot \nabla_{\bar{p}_{x|z}} L\left(\bar p_{x|z=0}^{(t)},\bar p^{(t)}_{x|z=1}, \lambda^{(t)};\theta^{(t)}\right)$
        \STATE \textit{Project $ \bar{p}_{x|z}^{(t+1)}$ onto $\ell_1$-norm constraints}: $||\bar{p}_{x|z}^{(t+1)} - p_{\tilde{x}|z}||_1 \leq 2\gamma_z, || \bar{p}_{x|z}^{(t+1)} ||_1 = 1$
\ENDFOR   
\ENDFOR
\STATE \textbf{return} { $\theta^{(t^{*})}$ and $\bar{p}_{x|z}^{(t^{*})}, z=0,1$ where $t^{*}$ denotes the \textit{best} iterate that satisfies the constraints in (\Cref{opt:general-dro-emp}) with the lowest objective.}
\end{algorithmic}
\end{algorithm}

%% file: sec/app_double_ml.tex
\section{Application to Double Machine Learning}\label{app:double-ml}

In this section, we demonstrate how the robust optimization approach
can also be applied to the double machine learning
estimator~\citep{chernozhukov2018double}. In particular, we use a
partially linear model (PLM) as in \citet{mackey2018orthogonal}.

Given i.i.d samples of $(X, Y, Z)$, double machine learning estimates
ATE as
\begin{align}\label{eq:doubleml-ate}
    \hat \tau = \frac{\E[(Y-f(X; \theta_0))Z]}{\E[Z^2]},
\end{align} where we posit a PLM as the outcome model, i.e. $Y|X, Z
\sim \mathcal{N}(f(X; \theta_0) + \theta_1 Z, \sigma^2)$, and
$\theta_0, \theta_1$ represent the parameters in the
model~\citep{mackey2018orthogonal}. We denote the parameters that
interact with $X$ and $Z$ as $\theta_0$ and $\theta_1$ respectively.
We further denote the set of all the parameters as $\theta =
(\theta_0, \theta_1)$, and the parameterized ATE estimator as $f(X;
\theta)$.

Given a feasible $\bar{p}_{x|y,z}$, the ATE estimator in
\Cref{eq:doubleml-ate} can be fully expressed in terms of $\theta$ and
$\bar p_{x|y,z}, z=0,1$:
\begin{align}\label{eq:doubleml-ate-param}
    \hat \tau (\bar{p}_{x|y,z}; \theta) =
    \frac{\E_{Y,Z}[\E_{p_{x|y,z}}[(Y-f(X; \theta_0))Z]]}{\E[Z^2]}.
\end{align}

Double machine learning uses the first half of the samples to fit the
model $f(X; \theta)$, and uses the second half of the samples to
estimate the expectation in \Cref{eq:doubleml-ate-param}. Assuming
$Y|X, Z \sim \mathcal{N}(f(X; \theta_0) + \theta_1 Z, \sigma^2)$,
fitting the parameters $\theta$ using MLE results in solving the least
square problem $\E[(Y-f(X; \theta_0) - \theta_1 Z)^2] =0$.
Differentiating w.r.t. $\theta$, we solve for the optimal $\theta$
such that
\begin{align}\label{eq:doubleml-grad-h}
   \left|\E\big[ \nabla_{\theta}  (Y-f(X; \theta_0) - \theta_1 Z)^2\big]\right| = 0.
\end{align}

This step gives us the optimal parameter $\theta^\ast$ in
\Cref{eq:MLE-grad-h}. Therefore, using double machine learning, the
robust optimization problem for estimating ATE with noisy covariates
is in the same form as \Cref{eq:ate-general-param,eq:MLE-grad-h}, with
$\hat \tau$ and the constraint derived in \Cref{eq:doubleml-ate-param}
and \Cref{eq:doubleml-grad-h}.


%% file: sec/app_exp.tex
\section{Additional Experimental Details}\label{app:exp}

This section includes further details on the experimental setup, including the model training details and the hyper-parameters tuned. All code will be made available on GitHub.

\subsection{Synthetic data generation details}\label{app:synthetic-data-generation}

In this section, we include the full data generation details for the synthetic data settings.
 
\paragraph{Backdoor adjustment and IPW:} 

With the backdoor adjustment and IPW estimators, we synthetically generate two datasets with $X$ as the confounder.

First, to demonstrate the variability of the estimated ATE intervals, we synthetically generate a dataset with binary outcomes according to a logistic model. The data generation is as follows: we randomly generate five covariates, i.e. $X_i \stackrel{iid}{\sim} N(1, I_5)$. Then, we generate the treatment $Z_i$ from a logistic model where $P(Z_i=1) = \logit(\alpha_0+\alpha_1^\top X_i)$. We generate the binary outcome $Y_i$ from a logistic outcome model, where $P(Y_i=1) = \logit(\beta_0 + \beta_1 Z_i + \beta_2^\top X_i)$. To illustrate the impact of each covariates, the coefficients of the logistic model are randomly drawn from a grid $\{-1,1\}$. As a result, we used $\alpha_0 = -1, \alpha_1 = (1, -1, 1,1, -1)^\top$, $\beta_0 = -1, \beta_1 = 1, \beta_2 = (-1, -1, -1, 1,1)^\top$. 

We further consider another synthetic dataset with a more complicated nonlinear outcome model and continuous outcomes. We use the Kang and Schafer example \citep{kang2007demystifying}, which consists of four
unobserved covariates $U_i \stackrel{iid}{\sim} N(0, I_4)$, $ i= 1,..., n$. They are used to generate four observed covariates $X_i$:
$X_{i1} = \exp(U_{i1}/2),$ $X_{i2} = U_{i2}/\{{1 + \exp(U_{i1})}\}+ 10,$
$X_{i3} = (U_{i1}U_{i3}+0.6)^3,$ and $X_{i4} = (U_{i2}+U_{i4}+20)^2.$
The outcome variable $Y_i$ is generated by $Y_i = 210 +
27.4U_{i1} + 13.7 2U_{i2} + 13.7U_{i3} +13.7U_{i4}+\epsilon_i$ where
$\epsilon_i\stackrel{iid}{\sim} N(0, 1)$. The treatment $Z_i$ is generated as a Bernoulli random variable with $P(Z_i=1) = \exp(-U_{i1} - 2U_{i2} - 0.25U_{i3} - 0.1U_{i4})$. 

\paragraph{Frontdoor adjustment:} For front adjustment, we synthetically generate two datasets, with $X$ denoting the mediators. First, we generate a dataset with binary outcomes using a logistic model with a single mediator. We begin with generating a noiseless confounder $U_i \in \br^2$ which is randomly drawn from a multivariate Gaussian distribution, i.e. $U_i \stackrel{iid}{\sim} N(1, I_5)$. Then, we generate the treatment $Z_i$ from a logistic model where $P(Z_i=1) = \logit(a_0+a_1^\top U_i)$. We then generate a binary mediator $X$ following a logistic model as well, i.e. $P(X_i=1) = \logit(\gamma_0+\gamma_1^\top Z_i)$. Lastly, the binary outcome $Y_i$ is generated as: $P(Y_i=1) = \logit(\beta_0 + \beta_1 X_i + \beta_2^\top U_i)$. The values of the coefficients are drawn from a grid. We used $a_0 = -1, a_1 = (1, -1, 1,1, -1)^\top$ for generating the treatment; and $\beta_0 = 1,\beta_1 = -1, \beta_2=(-1, -1, -1, 1,1)^\top$ for generating the outcome; $\gamma_0 = \gamma_1= 1$ for generating the mediator $X$.

Next, we generate another more complicated using a similar data generation as in \citet{jung2020estimating}. This data generation is more complicated with multiple mediators. In this model, the unobserved confounder $U_i$ is generated as $U_i \stackrel{iid}{\sim} N(-2, 1)$. Then, we generate the treatment $Z_i$ by drawing from a Bernoulli distribution with $P(Z_i = 1) = \logit(U_i + \epsilon_z)$, where $\epsilon_z \sim N(0,0.5)$. We further generate five covariates as the mediators, i.e. $X \in \br^5$. For each entry of $X$, it is drawn from a Bernoulli distribution with $P(X[i] = 1) = \logit(c_1 + c_2 * Z_i + \epsilon_x)$, where $\epsilon_x \sim N(-1,1)$, and the coefficients $c_1, c_2$ are drawn independently from  a Gaussian distribution $N(-2, 1)$. Lastly, the outcome is generated as: $Y_i \sim \text{Bern}( \logit(2 \beta^\top X_i + U_i + \epsilon_y)$, where $\epsilon_y \sim N(-1,1)$, and $\beta \sim N(1,1)$.

\paragraph{Generating noisy covariates:} 
Given the true covariates, we generate noisy covariates by synthetically adding a small amount of random noise to the noiseless covariates. In this way, we have both the access to the ground truth covariates and the noisy covariates. The the ground truth covariates enables us to estimate the true ATE. For each selected example, we perturb it by adding a noise that is drawn from a Gaussian distribution to each dimension. We then evaluate the performance of the different algorithms ranging from small to large amounts of noise. In this way, for backdoor adjustment and IPW with the binary outcome dataset, we select three levels of noise with mean = 0.1/0.3/0.5 and standard deviation = 0.5/0.5/1. The same noise levels are used in the dataset for frontdoor adjustment with a single mediator. For the Kang and Schafer example, we select five levels of Gaussian noise with mean = 1/2/3/4/5 and standard deviation at 1. For frontdoor adjustment with the multi-mediator dataset, we generated five levels of random Gaussian noise with mean = 0.1/0.2/0.3/0.4/0.5 and standard deviations at 0.5/0.5/1/1/1. We heuristically select the TV upper bound for each noise level as 0.1/0.2/0.3/0.4/0.5.

\subsection{Optimization code details}
For all the simulation studies and real case studies, we performed experiments comparing the na{\"i}ve approach and the robust causal approach.  All optimization code was written in Python and TensorFlow.\footnote{Abadi, M. et al. TensorFlow: Large-scale machine learning on heterogeneous systems,
2015. tensorflow.org.} All gradient steps were implemented using TensorFlow's Adam optimizer.
The experiments can also be reproduced using simple gradient descent without momentum. We computed full gradients over all datasets, but minibatching can also be used for very large datasets. Implementations for all approaches are included in the attached code. Training time was less than 10 minutes per model on a Laptop with a 2.3 GHz 8-Core Intel Core i9 CPU.

In the experiments, we replace all expectations in the objective and constraints with finite-sample empirical versions. For the naive approach with the backdoor and IPW adjustments, we used the DoWhy Library~\citep{dowhy}, which is publicly available online. Specifically, gor backdoor adjustment, we used \texttt{backdoor.linear\_regression}; for IPW, we used \texttt{backdoor.propensity\_score\_weighting}, which are both implemented in the DoWhy library. For RCI-NC and the naive approach with the frontdoor adjustment, we used a linear model, i.e. $f(x;\theta)=x^\top \theta$. Additionally, for the Kang and Schafer example, we standardized the dataset such that each covariate has zero mean and unit variance.

\begin{table}[!ht]
\caption{Hyperparameters for each approach}
\label{table: hparams}
\vskip 0.15in
\begin{center}
\begin{small}
\begin{sc}
\begin{tabular}{llll}
\toprule
Hparam & Values tried & Relevant approaches & Description \\
\midrule
$\eta_\theta$ & \{0.0001, 0.001,0.01,0.1\} & RCI-NC; Naive (frontdoor) & learning rate for $\theta$ \\
$\eta_\lambda$ & \{0.5,1.0,2.0\} & RCI-NC& learning rate for $\lambda$\\
$\eta_{z}, z \in\{0,1\}$    & \{0.001, 0.005, 0.01, 0.1\} & RCI-NC & learning rate for $\tilde{p}_z$ \\
\bottomrule
\end{tabular}
\end{sc}
\end{small}
\end{center}
\vskip -0.1in
\end{table}

\subsection{Hyperparameters and runtime details}
The hyperparameters for each approach were chosen to achieve the best performance on the coverage probability, where ``best'' is defined as the set of hyperparameters that achieved the highest coverage probability while satisfying all constraints relevant to the approach. The final hyperparameter values selected for each method were neither the largest nor smallest of all values tried. A list of all hyperparameters tuned and the values tried is given in Table \ref{table: hparams}.  

For both Table 1 and 3 simulations, each trial for CEVAE takes around ten minutes on an Nvidia GeForce GPU. We run for five noise levels, each level has 100 pairs of clean data and noise data. For the robust algorithm, it is much more lightweight and each trial takes around 25 minutes on a 2.3 GHz 8-Core Intel Core i9 laptop.

\subsection{Further case studies dataset details}\label{app:real-data-details}

\paragraph{ACIC dataset.}

We generate our dataset directly using ACIC's data generating process. The ACIC competition does not use the original data from UCI directly, but instead generates modified versions using pre-specified data generating processes with the known true ATE. We specifically use their ``modification 1'' out of four different modifications of the spam email dataset, for which code is also available on the ACIC 2019 website. 

Given the true covariates, we further generate noisy covariates by synthetically adding a small amount of noise at random, using a similar procedure as for the synthetic data. Specifically, we generated five levels of Gaussian noise with mean = 0.1/0.2/0.3/0.4/0.5 and standard deviations at 0.5/0.5/1/1/1. 

\paragraph{IHDP dataset.}

For the IHDP dataset, we used a procedure similar to \citet{hill2011bayesian, gupta2020estimating} to simulate the mediator and the outcome with the covariates and the treatment assignment from. The mediator $X$ takes the form $X \sim \mathcal{N}(c Z, \sigma^2_{u_m})$, where $Z$ is the treatment. The outcome $Y$ takes the form $Y \sim \mathcal{N}(a X + W\mathbf{b}, 1)$ where $W$ is the matrix of standardized (zero mean and unit variance) covariates 
and values in the vector $\mathbf{b}$ are randomly sampled
(0, 1, 2, 3, 4) with probabilities (0.5, 0.2, 0.15, 0.1, 0.05). The ground truth causal effect is $c \times a$. As a setting shown in~\citet{gupta2020estimating} where frontdoor adjustment outperforms backdoor adjust, we used $a=10,c=1,\sigma_{u_m} = 2$. 

\subsection{Additional experimental results}\label{app-exp-results}

\begin{figure*}[!ht]
  \begin{center}
    \begin{tabular}{ccc}
           \includegraphics[width=0.27\textwidth]{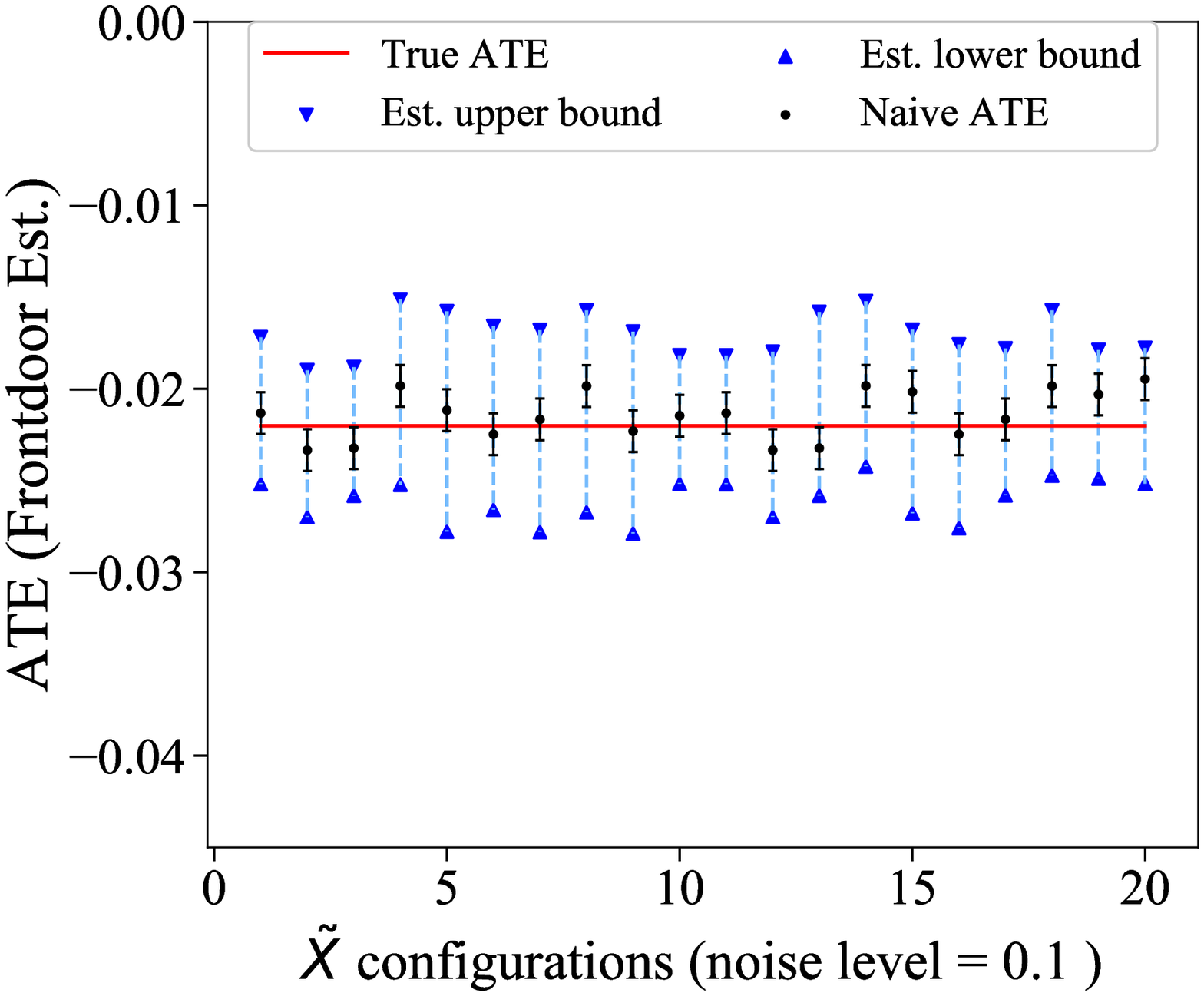} &\includegraphics[width=0.27\textwidth]{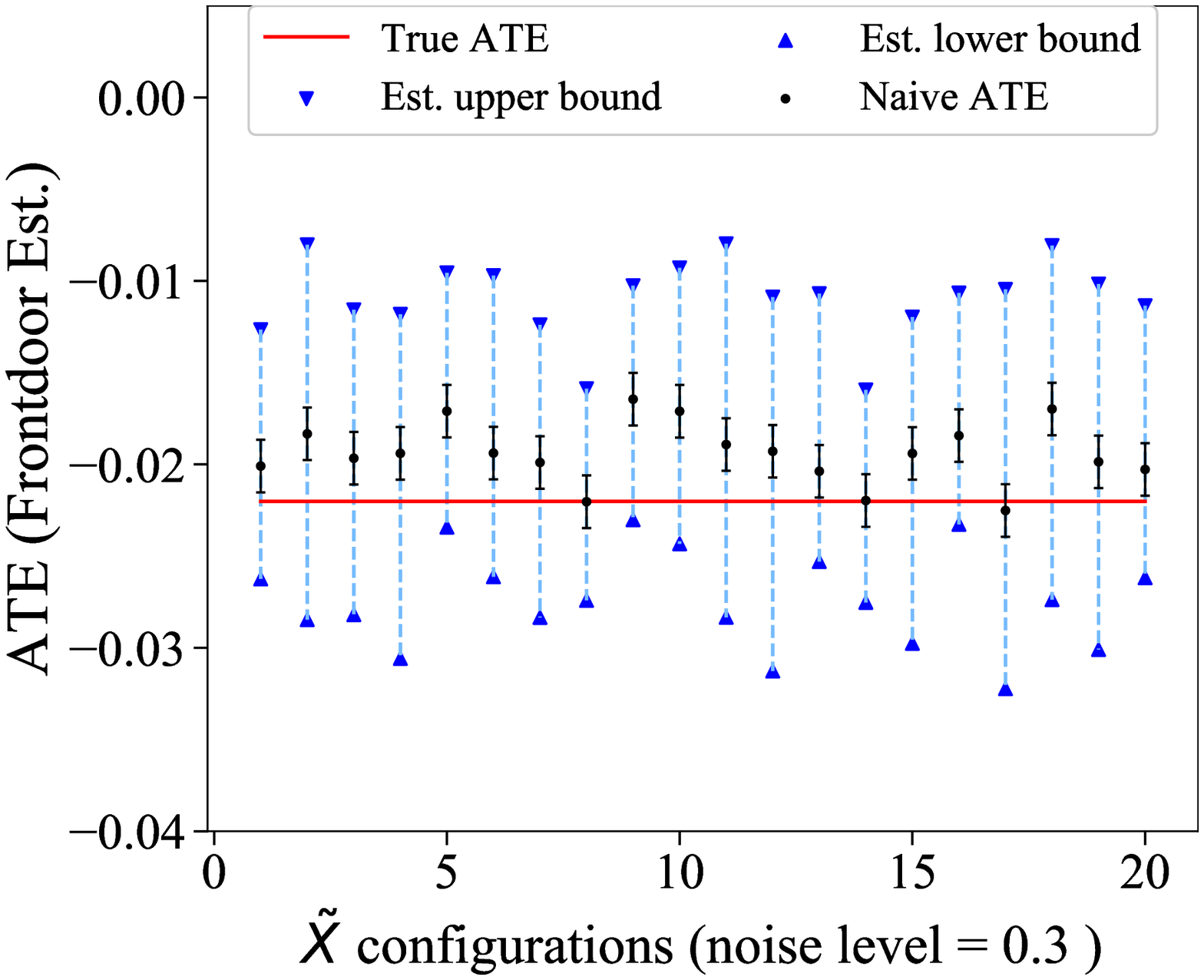} & \includegraphics[width=0.27\textwidth]{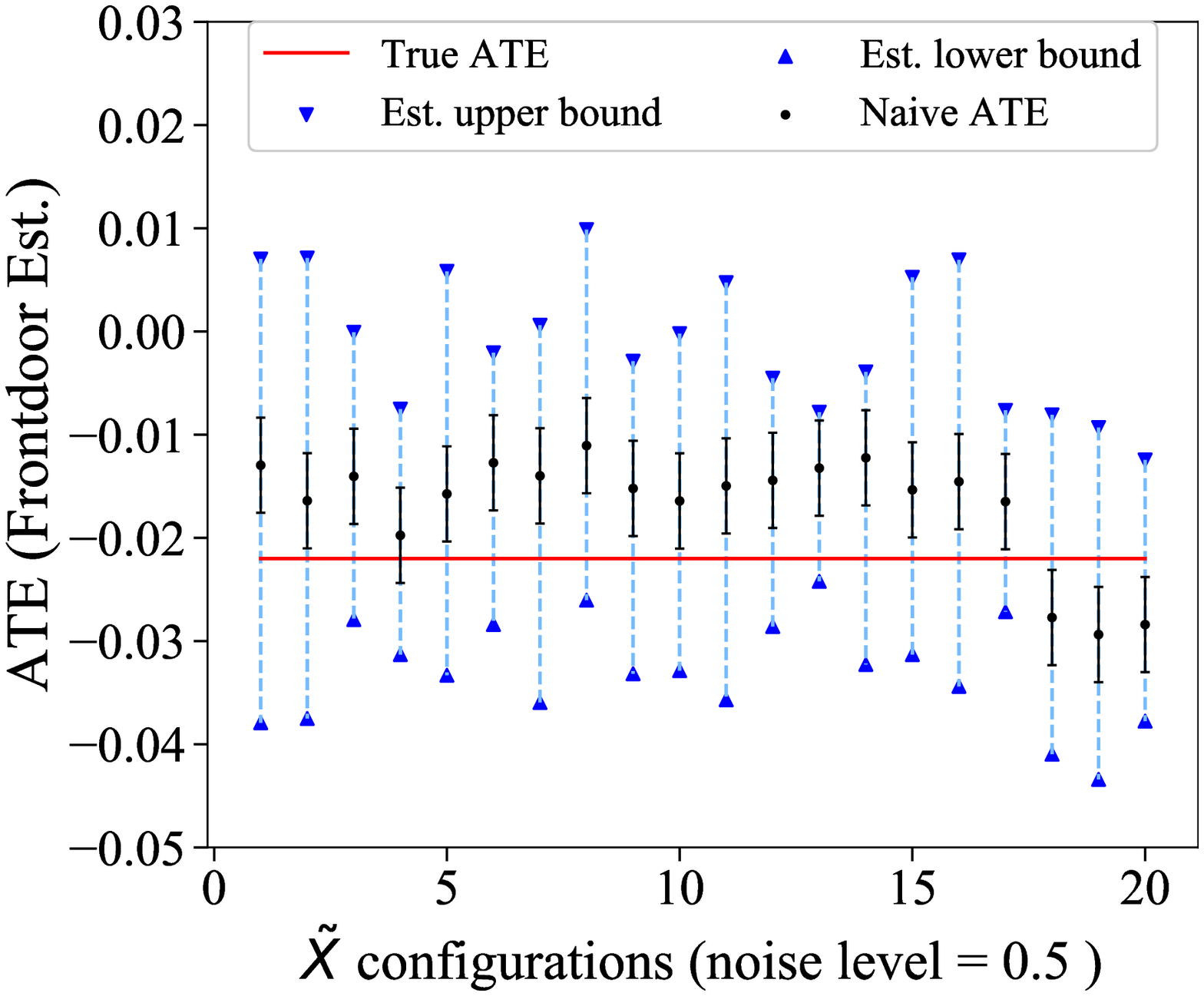} \\
          \; (a) & \;\; (b) & \;\;\; (c) 
    \end{tabular}
    \caption{Partial identification of ATE with  frontdoor adjustment estimators on synthetic dataset with binary outcome. The three noise levels have random Gaussian noise with mean = 0.1/0.3/0.5 and standard deviation = 0.5/0.5/1. In all plots, we compare RCI (this work) to the naive approach. The error bars indicate 95\% confidence interval of the naive ATE estimation over twenty trials. \textit{Intervals covering the true ATE is better.}} 
    \label{fig:frontdoor-sim}
  \end{center}
\end{figure*}

We include the plot of the ATE intervals obtained by RCI and the true ATE, and the results of the naive ATE estimator for the frontdoor adjustment estimator in Figure~\ref{fig:frontdoor-sim}.

